%% file: main.tex
\documentclass{article}
\usepackage{spconf,amsmath,graphicx}
\usepackage{url}

\title{End-to-end Neural Coreference Resolution Revisited: A Simple yet Effective Baseline}
\name{Tuan Manh Lai\,\textsuperscript{1} \qquad Trung Bui\,\textsuperscript{2} \qquad Doo Soon Kim\,\textsuperscript{3}}
\address{\textsuperscript{1} University of Illinois at Urbana-Champaign, USA \\
		 \textsuperscript{2} Adobe Research, San Jose, CA \\
		 \textsuperscript{3} Roku Inc., San Jose, CA}
\begin{document}
%
\maketitle
\begin{abstract}
Since the first end-to-end neural coreference resolution model was introduced, many extensions to the model have been proposed, ranging from using higher-order inference to directly optimizing evaluation metrics using reinforcement learning. Despite improving the coreference resolution performance by a large margin, these extensions add substantial extra complexity to the original model. Motivated by this observation and the recent advances in pre-trained Transformer language models, we propose a simple yet effective baseline for coreference resolution. Even though our model is a simplified version of the original neural coreference resolution model, it achieves impressive performance, outperforming all recent extended works on the public English OntoNotes benchmark. Our work provides evidence for the necessity of carefully justifying the complexity of existing or newly proposed models, as introducing a conceptual or practical simplification to an existing model can still yield competitive results.
\end{abstract}
\section{Introduction}
\input{introduction.tex}

\section{Method}\label{sec:method}
\input{methods.tex}

\section{Experiments and Results}\label{sec:experiments_and_results}
\input{experiments_and_results.tex}

\section{Conclusions}\label{sec:conclusions}
\input{conclusion.tex}

\bibliographystyle{IEEEbib}
\bibliography{strings,refs}

\end{document}

%% file: introduction.tex
\begin{figure}[ht!]
\centering
\includegraphics[width=0.45\textwidth]{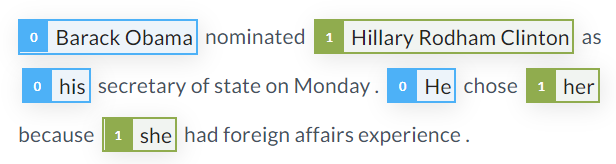}
\caption{An example of coreference resolution. There are two coreference chains in this example.}
\label{fig:coref_example}
\end{figure}

Coreference resolution is the task of clustering mentions in text that refer to the same entities \cite{ng2010review} (Figure \ref{fig:coref_example}). As a fundamental task of natural language processing, coreference resolution can be an essential component for many downstream applications. Many traditional coreference resolution systems are pipelined systems, each consists of two separate components: (1) a mention detector for identifying entity mentions from text (2) a coreference resolver for clustering the extracted mentions \cite{raghunathanEtal2010multi,durrettklein2013easy,clarkmanning2015entity,wisemanetal2016learning,clarkmanning2016deep}. These models typically rely heavily on syntatic parsers and use highly engineered mention proposal algorithms.

\begin{figure}[ht!]
\centering
\includegraphics[width=0.5\textwidth]{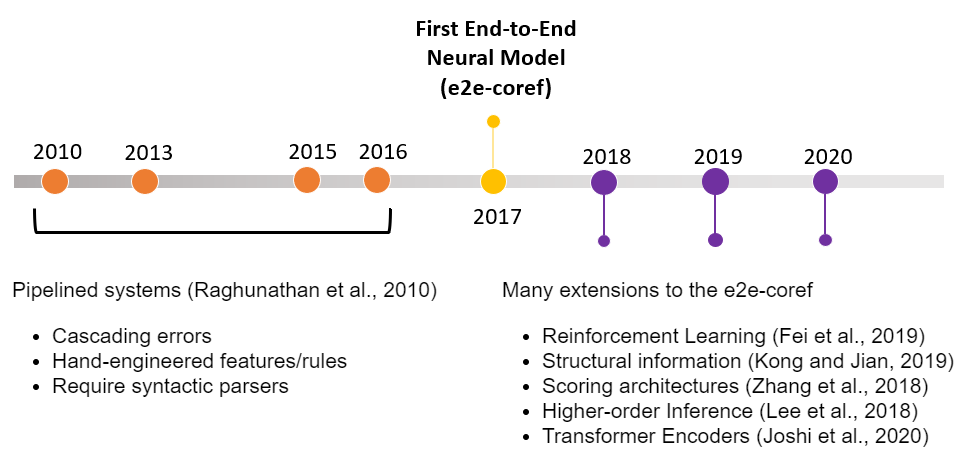}
\caption{An overview of coreference resolution research in the last decade. Pipelined systems were heavily used before the introduction of \texttt{e2e-coref}. Since 2017, various extensions to the model have been proposed.}
\label{fig:coref_timeline}
\end{figure}

In 2017, the first end-to-end coreference resolution model named \texttt{e2e-coref} was proposed \cite{leeetal2017end}. It outperforms previous pipelined systems without using any syntactic parser or complicated hand-engineered features. Since then, many extensions to the \texttt{e2e-coref} model have been introduced, ranging from using higher-order inference to directly optimizing evaluation metrics using reinforcement learning \cite{zhangetal2018neuralcoreference,leeetal2018higher,Gu2018ASO,feietal2019end,kantorgloberson2019coreference,joshietal2019bert,ijcai20190700,joshi2020spanbert} (Figure \ref{fig:coref_timeline}). Despite improving the coreference resolution performance by a large margin, these extensions add a lot of extra complexity to the original model. Motivated by this observation and the recent advances in pre-trained Transformer language models, we propose a simple yet effective baseline for coreference resolution. We introduce simplifications to the original \texttt{e2e-coref} model, creating a conceptually simpler model for coreference resolution. Despite its simplicity, our model outperforms all aforementioned methods on the public English OntoNotes benchmark. Our work provides evidence for the necessity of carefully justifying the complexity of existing or newly proposed models, as introducing a conceptual or practical simplification to an existing model can still yield competitive results. The findings of our work agree with the results of several recent studies \cite{xuchoi2020revealing,Kirstain2021CoreferenceRW,lai2021bert}.


%% file: methods.tex
\begin{figure}[!t]
    \centering
    \includegraphics[width=0.45\textwidth]{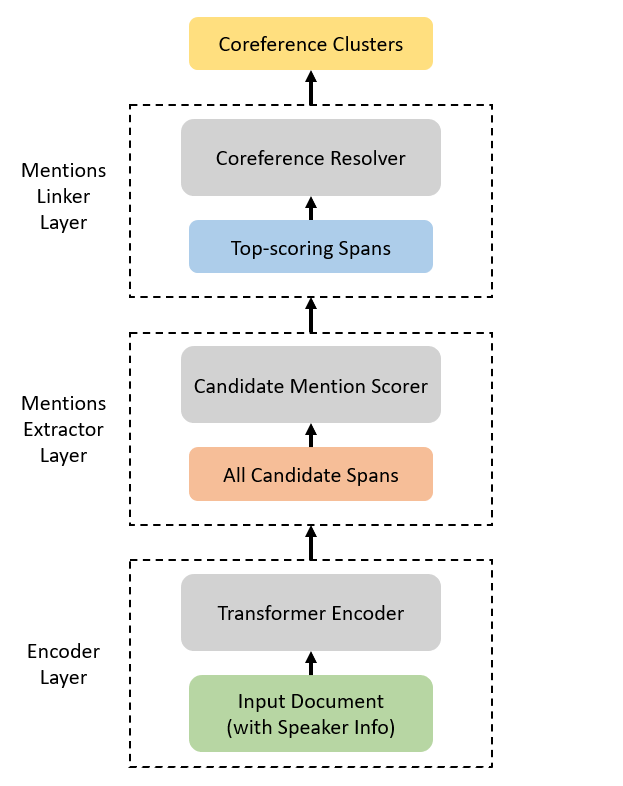}
    \caption{A high level overview of our proposed model for coreference resolution.}
    \label{fig:high_level}
\end{figure}

At a high level, our coreference resolution model is similar to the \texttt{e2e-coref} model (Figure \ref{fig:high_level}). Given a sequence of tokens from an input document, the model first forms a contextualized representation for each token using a Transformer-based encoder. After that, all the spans (up to a certain length) in the document are enumerated. The model then assigns a score to each candidate span indicating whether the span is an entity mention. A portion of top-scoring spans is extracted and fed to the next stage where the model predicts distributions over possible antecedents for each extracted span. The final coreference clusters can be naturally constructed from the antecedent predictions. In the following subsections, we go into more specific details.

\input{methods_notations_and_preliminaries}

\input{methods_encoder_layer}

\input{methods_mention_extractor_layer}

\input{methods_mention_linker_layer}

%% file: methods_notations_and_preliminaries.tex
\begin{figure}[!t]
    \centering
    \includegraphics[width=0.45\textwidth]{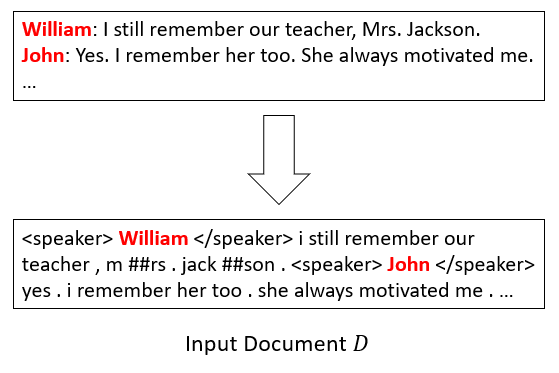}
    \caption{An example illustrating the strategy of concatenating the speaker’s name with the corresponding utterance (assuming the model utilizes WordPiece for tokenization).}
    \label{fig:speaker_info_example}
\end{figure}

\subsection{Notations and Preliminaries}\label{subsection:notations_and_prelim}
Given an input document $D = (t_1, t_2, ..., t_n)$ consisting of $n$ tokens, the total number of possible text spans is  $N = n(n+1)/2$. For each span $i$,  we denote the start and end indices of the span by $\texttt{START}(i)$ and $\texttt{END}(i)$ respectively. We also assume an ordering of the spans based on $\texttt{START}(i)$; spans with the same start index are ordered by $\texttt{END}(i)$. Furthermore, we only consider spans that are entirely within a sentence and limit spans to a max length of $L$.

Since the speaker information is known to contain useful information for coreference resolution, it has been extensively used in previous works \cite{durrettklein2013easy,leeetal2017end,leeetal2018higher,joshietal2019bert,joshi2020spanbert}. For example, the original \texttt{e2e-coref} model converts speaker information into binary features indicating whether two candidate mentions are from the same speaker. In this work, we employ a more intuitive strategy that directly concatenates the speaker’s name with the corresponding utterance \cite{wuetal2020corefqa}. This straightforward strategy is simple to implement and has been shown to be more effective than the feature-based method \cite{wuetal2020corefqa}. Figure \ref{fig:speaker_info_example} illustrates the concatenation strategy.

%% file: methods_encoder_layer.tex
\subsection{Encoder Layer}
Given the input document $D = (t_1, t_2, ..., t_n)$, the model simply forms a contextualized representation for each input token, using a Transformer-based encoder such as BERT \cite{devlinetal2019bert} or SpanBERT \cite{joshi2020spanbert}. These pretrained language models typically can only run on sequences with at most 512 tokens. Therefore, to encode a long document (i.e., $n > 512$), we split the document into overlapping segments by creating a $n$-sized segment after every $n/2$ tokens. These segments are then passed on to the Transformer-based encoder independently. The final token representations are derived by taking the token representations with maximum context. Let $\textbf{X} = (\textbf{x}_1, \textbf{x}_2, ..., \textbf{x}_n)$ be the output of the Transformer encoder.

Note that the \texttt{e2e-coref} model uses the GloVe and Turian embeddings \cite{turianetal2010word,penningtonetal2014glove} and character embeddings produced by 1-dimensional convolution neural networks. From an implementation point of view, it is easier to use a Transformer encoder than combining these traditional embeddings. For example, the \texttt{Transformers} library\footnote{\url{https://github.com/huggingface/transformers}} allows users to experiment with various state-of-the-art Transformer-based models by simply writing few lines of code.

Now, for each span $i$, its span representation $\textbf{g}_i$ is defined as:
\begin{equation}
    \textbf{g}_i = \big\lbrack\textbf{x}_{\texttt{START}(i)}, \textbf{x}_{\texttt{END}(i)}, \hat{\textbf{x}}_{i} \big\rbrack
\end{equation}
where $\textbf{x}_{\texttt{START}(i)}$ and $\textbf{x}_{\texttt{END}(i)}$ are the boundary representations, consisting of the first and the last token representations of the span $i$. And $\hat{\textbf{x}}_{i}$ is computed using an attention mechanism \cite{Bahdanau2015NeuralMT} as follows:
\begin{equation}
\begin{split}
\alpha_t & = \text{FFNN}_{\alpha}(\textbf{x}_t) \\
\beta_{i,t} & = \frac{\exp{(\alpha_t)}}{\sum \limits_{j=\texttt{START}(i)}^{\texttt{END}(i)} \exp{(\alpha_j)}} \\
\hat{\textbf{x}}_{i} &= \sum \limits_{j=\texttt{START}(i)}^{\texttt{END}(i)} \beta_{i,j} \,\textbf{x}_j
\end{split}
\end{equation}
where $\text{FFNN}_{\alpha}$ is a multi-layer feedforward neural network that maps each token-level representation $\textbf{x}_t$ into an unnormalized attention score. $\hat{\textbf{x}}_{i}$ is a weighted sum of token vectors in the span $i$. Our span representation generation process closely follows that of \texttt{e2e-coref}. However, a simplification we make is that we do not include any additional features such as the size of span $i$ in its representation $\textbf{g}_i$.

%% file: methods_mention_extractor_layer.tex
\subsection{Mentions Extractor Layer}\label{subsec:mention_extractor}
In this layer, we first enumerate all the spans (up to a certain length $L$) in the document. For each span $i$, we simply use a feedforward neural network $\text{FFNN}_\text{m}$ to compute its mention score:
\begin{equation}
\label{eqn:mention_score}
    s_m(i) = \text{FFNN}_\text{m}(\textbf{g}_i)
\end{equation}
After this step, we only keep up to $\lambda n$ spans with the highest mention scores. In previous works, to maintain a high recall of gold mentions, $\lambda$ is typically set to be $0.4$ \cite{leeetal2017end,leeetal2018higher}. These works do not directly train the mention extractor: The mention extractor and the mention linker are jointly trained to only maximize the marginal likelihood of gold antecedent spans.

In coreference resolution datasets such as the OntoNotes benchmark \cite{Pradhan2012CoNLL2012ST}, singleton mentions are not explicitly labeled, because the annotations contain only mentions that belong to a coreference chain. However, these annotations of non-singleton mentions can still provide useful signals for training an efficient mention extractor \cite{zhangetal2018neuralcoreference}. Thus, we also propose to pre-train our mention extractor using these annotations. In Section \ref{sec:experiments_and_results},  we will empirically demonstrate that this pre-training step greatly improves the performance of our mention extractor layer. As a result, we only need to set the parameter $\lambda$ to be 0.25 in order to maintain a high recall of gold mentions. To this end, the pretraining loss is calculated as follows:
\begin{equation}
\label{equation:loss_detect}
\begin{split}
\mathcal{L}_{\text{detect}}(i) &= y_i \log{\hat{y}_i} + (1-y_i) \log{(1-\hat{y}_i)} \\
\mathcal{L}_{\text{detect}} &= - \sum_{i\,\in\,\text{S}} \mathcal{L}_{\text{detect}}(i)
\end{split}
\end{equation}
where $\hat{y}_i = \text{sigmoid}(s_m(i))$, and $y_i = 1$ if and only if the span $i$ is a mention in one of the coreference chains. $S$ is the set of the top scoring spans (and so $|S| \leq \lambda n$).

%% file: methods_mention_linker_layer.tex
\subsection{Mentions Linker Layer}
For each span $i$ extracted by the mention extractor, the mention linker needs to assign an antecedent $a_i$ from all preceding spans or a dummy antecedent $\epsilon$: $a_i \in Y(i) = \{\epsilon, 1, \dots, i-1\}$ (the ordering of spans was discussed in Subsection \ref{subsection:notations_and_prelim}). The dummy antecedent $\epsilon$ represents two possible cases. One case is the span itself is not an entity mention. The other case is the span is an entity mention but it is not coreferent with any previous span extracted by the mention extractor.

The coreference score $s(i,j)$ of two spans $i$ and $j$ is computed as follows:
\begin{equation}
\begin{split}
s_a(i, j) &= \text{FFNN}_\text{a}\big(\big\lbrack \textbf{g}_i, \textbf{g}_j, \textbf{g}_i \circ \textbf{g}_j \big\rbrack\big) \\
s(i, j) &= s_m(i) + s_m(j) + s_a(i,j)
\end{split}
\end{equation}
where $\text{FFNN}_\text{a}$ is a feedforward network. $s_m(i)$ and $s_m(j)$ are calculated using Equation \ref{eqn:mention_score}. The score $s(i,j)$ is affected by three factors: (1) $s_m(i)$, whether span $i$ is a mention, (2) $s_m(j)$, whether span $j$ is a mention, and (3) $s_a(i,j)$ whether $j$ is an antecedent of $i$. In the special case of the dummy antecedent, $s(i, \epsilon)$ is fixed to 0. In the \texttt{e2e-coref} model, when computing $s_a(i,j)$, a vector encoding additional features such as genre information and the distance between the two spans is also used. We do not use such a feature vector when computing $s_a(i,j)$ to simplify the implementation.

We want to maximize the marginal log-likelihood of all antecedents in the correct coreference chain for each mention:
\begin{equation}
\label{equation:loss_linker}
\log{\prod_{i \in S}\;{\sum_{\hat{y}\in Y(i) \cap \text{GOLD}(i)} P(\hat{y})}}
\end{equation}
where $S$ is the set of the top scoring spans extracted by the mention extractor (i.e., the set of unpruned spans). $\text{GOLD}(i)$ is the set of spans in the gold cluster containing span $i$. If span $i$ does not belong to any coreference chain or all gold antecedents have been pruned, then $\text{GOLD}(i) = \{\epsilon\}$.

\input{tables/overall_result}

\input{tables/mention_recall}

To summarize, we first pre-train the mention extractor to minimize the loss function defined in Eq. \ref{equation:loss_detect}. We then jointly train the mention extractor and the mention linker to optimize the objective defined in Eq. \ref{equation:loss_linker} in an end-to-end manner.

%% file: tables/overall_result.tex
\begin{table*}[ht!]
\tiny
\centering
\resizebox{\textwidth}{!}{%
\begin{tabular}{l|ccc|ccc|ccc|c}
\hline
 & \multicolumn{3}{c|}{MUC} & \multicolumn{3}{c|}{B-CUBED} & \multicolumn{3}{c|}{$\text{CEAF}_{\phi_4}$} & \multicolumn{1}{l}{} \\
 & P & R & F1 & P & R & F1 & P & R & F1 & \multicolumn{1}{l}{Avg. F1} \\ \hline
e2e-coref \cite{leeetal2017end} & 78.4 & 73.4 & 75.8 & 68.6 & 61.8 & 65.0 & 62.7 & 59.0 & 60.8 & 67.2 \\
e2e-coref + Structural info \cite{ijcai20190700} & 80.5 & 73.9 & 77.1 & 71.2 & 61.5 & 66.0 & 64.3 & 61.1 & 62.7 & 68.6 \\
c2f-coref + ELMo \cite{leeetal2018higher} & 81.4 & 79.5 & 80.4 & 72.2 & 69.5 & 70.8 & 68.2 & 67.1 & 67.6 & 73.0 \\
EE + BERT-large \cite{kantorgloberson2019coreference} & 82.6 & 84.1 & 83.4 & 73.3 & 76.2 & 74.7 & 72.4 & 71.1 & 71.8 & 76.6 \\
c2f-coref + BERT-large \cite{joshietal2019bert} & 84.7 & 82.4 & 83.5 & 76.5 & 74.0 & 75.3 & 74.1 & 69.8 & 71.9 & 76.9 \\
c2f-coref + SpanBERT-large \cite{joshi2020spanbert} & \textbf{85.8} & 84.8 & 85.3 & 78.3 & 77.9 & 78.1 & \textbf{76.4} & \textbf{74.2} & \textbf{75.3} & 79.6 \\ \hline
Simplified e2e-coref (Ours) & {85.4} & \textbf{85.4} & \textbf{85.4} & \textbf{78.4} & \textbf{78.9} & \textbf{78.7} & 76.1 & 73.9 & 75.0 & \textbf{79.7} \\ \hline
\end{tabular}%
}
\caption{Performance on the OntoNotes coreference resolution benchmark.}
\label{tab:overall_result}
\end{table*}

%% file: tables/mention_recall.tex
\begin{table}[ht!]
\centering
\resizebox{\linewidth}{!}{%
\begin{tabular}{|l|c|c|}
\hline
 & \multicolumn{1}{l|}{Avg. Nb Spans Proposed} & \multicolumn{1}{l|}{Gold Mention Recall} \\ \hline
e2e-coref \cite{leeetal2017end} & $\sim$ 200.43 spans / docs  & 92.7\% \\ \hline
Simplified e2e-coref (Ours) & $\sim$ \textbf{141.79 spans / docs} & \textbf{95.7\%} \\ \hline
\end{tabular}%
}
\caption{Proportion of gold mentions covered in the development data by the mention extractor of e2e-coref and our mention extractor.}
\label{tab:mention_recall}
\end{table}

%% file: experiments_and_results.tex
\textbf{Dataset and Experiments Setup} To evaluate the effectiveness of the proposed approach, we use the CoNLL-2012 Shared Task English data \cite{Pradhan2012CoNLL2012ST} which is based on the OntoNotes corpus. This dataset has 2802/343/348 documents for the train/dev/test split. Similar to previous works, we report precision, recall, and F1 of the MUC, $\text{B}^3$, and $\text{CEAF}_{\phi_4}$ metrics, and also average the F1 score of all three metrics. We used SpanBERT (spanbert-large-cased) \cite{joshi2020spanbert} as the encoder. Two different learning rates are used, one for the lower pretrained SpanBERT encoder (5e-05) and one for the upper layers (1e-4). We also use learning rate decay. The number of training epochs is set to be 100. The batch size is set to be 32. We did hyper-parameter tuning using the provided dev set. To train our model, we use two 16GB V100 GPUs and use techniques such as gradient checkpointing and gradient accumulation to avoid running out of GPUs' memory.

~\\\textbf{Comparison with Previous Methods} Table \ref{tab:overall_result} compares our model with several state-of-the-art coreference resolution systems. Overall, our model outperforms the original \texttt{e2e-coref} model and also all recent extended works. For example, compared to the variant [c2f-coref + SpanBERT-large], our model achieves higher F1-scores for the MUC and $\text{B}^3$ metrics. Even though our model achieves a slightly lower F1-score for the $\text{CEAF}_{\phi_4}$ metric, the overall averaged F1 score of our model is still better. The variant [c2f-coref + SpanBERT-large] is more complex than our method, because it has some other additional components such as coarse-to-fine antecedent pruning and higher-order inference \cite{leeetal2018higher,joshi2020spanbert}.

Recently, a model named \texttt{CorefQA} has been proposed \cite{wuetal2020corefqa}, and it achieves an averaged F1 score of $83.1$ on the English OntoNotes benchmark. The work takes a complete departure from the paradigm used by the \texttt{e2e-coref} model, and instead, proposes to formulate the coreference resolution problem as a span prediction task, like in question answering. To achieve its impressive performance, the \texttt{CorefQA} model is very computationally expensive. In order to predict coreference clusters for a single document, \texttt{CorefQA} needs to run a Transformer-based model on the same document many times (each time a different query is appended to the document).\\
~\\\textbf{Analysis on the Performance of the Mention Extractor} In our work, the value of the parameter $\lambda$ for pruning is set to be 0.25. On the other hand, it is set to be 0.4 in the \texttt{e2e-coref} model. Table \ref{tab:mention_recall} shows the comparison in more detail. Our mention extractor extracts 95.7\% of all the gold mentions in the dev set, while the mention extractor of \texttt{e2e-coref} extracts only 92.7\% of them. By proposing fewer candidate spans, the workload of our mention linker is also reduced.

%% file: conclusion.tex
In this work, we propose a simple yet effective baseline for the task of coreference resolution. Despite its simplicity, our model still outperforms all recent extended works on the English OntoNotes benchmark. In future work, we plan to reduce the computational complexity of our baseline model using compression techniques \cite{Sanh2019DistilBERTAD,sunetal2020mobilebert,lai2020simple}. We also plan to address the task of event coreference resolution \cite{laietal2021context,wen2021resin}.